\documentclass{article}

\usepackage{arxiv}

\usepackage[utf8]{inputenc} 
\usepackage[T1]{fontenc}    
\usepackage{hyperref}       
\usepackage{url}            
\usepackage{booktabs}       
\usepackage{amsfonts}       
\usepackage{nicefrac}       
\usepackage{microtype}      
\usepackage{lipsum}		
\usepackage{graphicx}
\usepackage{natbib}
\usepackage{doi}

\title{Formulating A Strategic Plan Based On Statistical Analyses And Applications For Financial Companies Through A Real-World Use Case}

\author{ \href{https://orcid.org/0000-0003-3314-4281}{\includegraphics[scale=0.06]{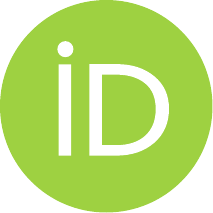}\hspace{1mm}Saman Sarraf}\thanks{Use footnote for providing further
		information about author (webpage, alternative
		address)---\emph{not} for acknowledging funding agencies.} \\
	The Institute of Electrical and Electronics Engineers\\
	Senior Member IEEE\\
	\texttt{samansarraf@ieee.org} \\
}

\renewcommand{\headeright}{Technical Report}
\renewcommand{\undertitle}{Technical Report}
\renewcommand{\shorttitle}{Strategic Plan Based On Statistical Analyses For Financial Companies}

\hypersetup{
pdftitle={Formulating A Strategic Plan Based On Statistical Analyses And Applications For Financial Companies Through A Real-World Use Case},
pdfsubject={Strategic Plan},
pdfauthor={Saman Sarraf},
pdfkeywords={Strategic Plan, Statistical Analysis, Financial Companies, Cloud Computing},
}

\begin{document}
\maketitle

\begin{abstract}
	Business statistics play a crucial role in implementing a data-driven strategic plan at the enterprise level to employ various analytics where the outcomes of such a plan enable an enterprise to enhance the decision-making process or to mitigate risks to the organization. In this work, a strategic plan informed by the statistical analysis is introduced for a financial company called LendingClub, where the plan is comprised of exploring the possibility of onboarding a big data platform along with advanced feature selection capacities. The main objectives of such a plan are to increase the company’s revenue while reducing the risks of granting loans to borrowers who cannot return their loans. In this study, different hypotheses formulated to address the company’s concerns are studied, where the results reveal that the amount of loans profoundly impacts the number of borrowers charging off their loans. Also, the proposed strategic plan includes onboarding advanced analytics such as machine learning technologies that allow the company to build better generalized data-driven predictive models.
\end{abstract}

\keywords{Strategic Plan, Statistical Analysis, Financial Companies, Cloud Computing}

\section{Introduction}
Formulating a strategic plan aligned with a company’s business scope allows the company to explore data-driven ways of business improvement and risk mitigation quantitively while utilizing collected data to perform statistical applications. The company’s business leadership generally organizes joint meetings with internal or external data analysis teams to design a plan for executing business-related statistical analysis. Such projects demonstrate that the company should invest in what areas and adjust the budget for business verticals with low revenue. Furthermore, statistical applications can determine the logic of how to improve staff performance in the workplace.
\\
LendingClub, as a peer-to-peer lending company, offers loans and investment products in different sectors, including personal and business loans, automobile loans, and health-related financing loans. LendingClub’s business model comprises three primary players: borrowers, investors, and portfolios for issued loans. 
LendingClub is about expanding the statistical analytics that consists of infrastructure and software algorithm applications to develop two meaningful solutions ultimately: a) estimating durations in which clients will pay off loans; and b) 30-minute loan approval decision-making. To implement these two capabilities, the company has collected data on loans that were granted or rejected over 12 years, including 145 attributes and more than 2 million observations, where 32 features have no missing values across the dataset.
\\
To achieve its ultimate targets, LendingClub performs a statistical analysis of numerous steps to determine whether to accept or reject hypotheses, which enables data scientists and statisticians to select attributes for predictive modeling. LendingClub seeks patterns in the loan data to discover relationships between a loan amount and borrowers who have charged off and reported by LendingClub \cite{emekter2015evaluating}. The company assumes a potential correlation between the two features, which establishes specific loan criteria for the group loan applicants who might encounter such an issue. Discovering the correlation enables LendingClub to enhance its risk management portfolio and minimize the risk of losing financial resources, aiming to mitigate the negative impacts of issuing loans to borrowers of this category. Using business statistics, the company seeks proof of concept for the mentioned ideas before recruiting a third-party software developer to implement a standalone product; therefore, the internal data scientists explore various aspects of such data, not limited to the questions listed above \cite{sarraf2016deepad,grady2016age}. 
\\
In the first phase, demographic information is extracted from the datasets, and data preprocessing steps, such as data cleaning, are performed to remove any broken data from the database. Next, further investigation of specific data (e.g., type of loans issued, loans issued by region, and a more in-depth analysis of bad loans) is performed \cite{sarraf2016classification,sarraf2016classificationMRI,sarraf2016deep}. In the second phase, which oversees the business perspective, the company’s experts explore the operative side of the business (operational business aspects) and analyze applicants’ income category. The third phase refers to the risk assessment of issuing loans, which consists of four steps: a) identifying existing risks in the business; b) the importance and role of credit scores in the loan approval or denial; c) defining bad loans and risky borrowers; d) loans by default (pre-approved); and e) exploring risks by targeted criteria \cite{sarraf20195g}. The ultimate goals of such extensive analysis are to lead LendingClub’s data scientists to explore the feasibility of answering the two questions above based on current data, provide recommendations for data collection, or modify the business scope \cite{saverino2016associative,sarraf2016big,sarraf2016robust}. 

\section{Problem Statement and Hypothesis}
\label{sec:headings}

The problem for this work points to statistical applications in LendingClub, which establishes three hypotheses regarding the relationship between the “Loan Amount” and “Charge OFF Flag” features, where various statistical analyses, including hypothesis testing \cite{bunting2019impact} and correlation analysis \cite{mondal2016sensitivity}, are employed. The hypotheses are as follows:
\begin{enumerate}
  \item Accepting or rejecting the hypothesis that any relationship exists between the loan amounts and charge-offs
  \item Accepting or rejecting the hypothesis that any relationship exists between the higher loan amounts and charge-offs
  \item Accepting or rejecting the hypothesis that any relationship exists between the lower loan amounts and charge-offs
\end{enumerate}
\section{Statistical Analysis Pipeline Design}
The problem statement consists of three main components: a) data exploration, b) descriptive analysis of loan duration, and c) real-time (fast) loan approval (or denial). Data exploration includes preprocessing, data cleaning, feature engineering, and selection to result in a meaningful descriptive analysis to find an accurate loan during and prediction. In the real-time step, various statistical techniques are explored, including hypothesis testing, student T-Test, and ANOVA testing, and statistical models, such as linear regression, logistic regression, cluster analysis, ANOVA tests, and correlation analysis \cite{anderson2017task,yang2018deep,strother2014hierarchy}.
\subsection{Data Exploration}
\begin{figure}
	\centering{\includegraphics[width=\textwidth,height=\textheight,keepaspectratio]{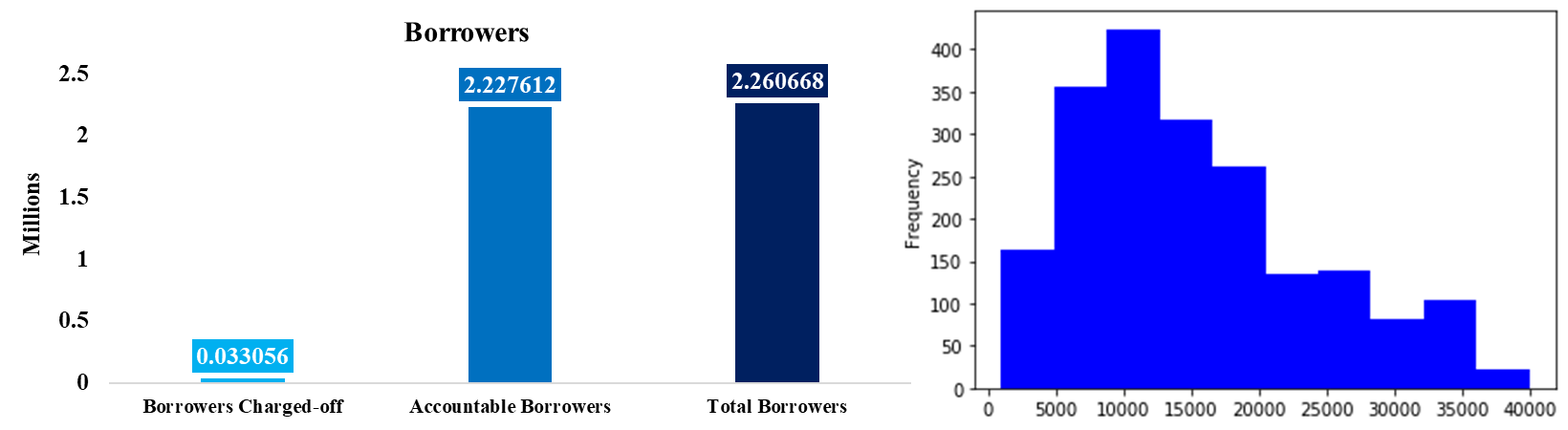}}
	
	\caption{Left: Distribution of borrowers Right: Distribution of loan amounts charged off}
	\label{fig:fig1}
\end{figure}

Missing values are removed from the loan data, and “loanAmnt” refers to “the listed amount of the loan applied for by the borrower if, at some point in time, the credit department reduces the loan amount, then it will be reflected in this value” and “debt\textunderscore settlement\textunderscore flag” indicating “flags whether or not the borrower, is charged-off, is working” are extracted from the preprocessed data shown in Figure \ref{fig:fig1}. The “debt\textunderscore settlement\textunderscore flag” – a binary feature – is considered a categorical attribute requiring conversion to numerical equivalents for statistical analysis \cite{vafaei2019normalization,sarraf2020binary}. Also, the histogram of loan amounts shows how borrowers are distributed regarding loan amounts.
\subsection{Hypothesis Testing}
In this experiment, T-Test is the primary method for whether to accept or reject the hypothesis. A T-Test is a hypothesis-testing method with broad applications in the industry due to its simplicity and convergence capability with a small sample of data \cite{staniewski2016contribution,sarraf2020recent}. T-Test requires a relatively small subset of data so that the loan dataset is shuffled, and a subsample of 1000 observations is randomly selected from charged-off samples along with 1000 samples, which are randomly selected from the on-time borrowers’ observations for further analysis\cite{sarraf2018french}. To explore the consistency of T-Test results, analysis of variance (ANOVA) tests are applied to the same subsets as those used in the previous method. ANOVA tests demonstrate whether such groups offer statistically significant differences \cite{quirk2016excel,sarraf2021comprehensive}.
\subsection{Correlation Analysis}
Correlation analysis is applied to the subsets to show the dependency between two features \cite{schober2018correlation}. This analysis can indicate whether the loan amount impacts the number of borrowers charged off. Correlation analysis provides additional exposure to the data, which might strengthen the acceptance or rejection of the three hypotheses\cite{sarraf2016hair,sarraf2017eeg}. 
\subsection{Results Visualization and Interpretation}
The results of statistical analysis methods are visualized and interpreted to verify whether the hypotheses are accepted. Also, the visualization of results allows the company’s data scientists to explore whether such outcomes from various techniques converge for decision-making and conclusion purposes.
\section{Summary of Results}
To perform an accurate T-Test, several data requirements must be met: a) test variables are continuous; b) test variables (observations) are independent; c) subsets are randomly selected; c) data distribution is approximately normal; d) variance scores of subsets and population are approximately consistent; and e) no outliers \cite{sarraf2014mathematical,sarraf2023ovitad,sarrafmultimodal}. In addition to these criteria, a balanced dataset design is required to conduct a meaningful ANOVA test, where the number of subjects in each group needs to be equal \cite{harmancioglu1999basic}. Also, an ideal correlation analysis requires data to be independently collected as paired samples, preferably continuous numeric values \cite{nerenberg1990correlation}. 
\subsection{Data Analysis}
The first step of data analysis is exploring the distribution of observations regarding the number of on-time borrowers versus those who have charged off. The next step is to downsample the charged-off samples into subsets of 1000 observations. The same procedure was applied to on-time borrowers’ observations (non-charged-off), and 1000 samples were randomly selected; thus, each subset included 2000 samples of each class equally distributed \cite{krishnan2016cognitive}. The mean, standard deviation, and variance of each subset were calculated. The statistical measures of subsets are highly similar, which suggests the need for statistical testing to produce interpretable results. Figure \ref{fig:fig1} shows a histogram of each subset where the number of bins is automatically calculated from the data (bin=10). The histogram results indicate that most of the issued loan amounts are in the range of [\$5000,\$20000].
\subsubsection{Hypothesis 1}
G*Power statistical software application \cite{faul2007g} performed a T-Test against each subset, including 2000 samples of charged-off and on-time borrowers’ observations equally distributed. One-tailed T-Tests were conducted using an alpha error probability of 0.05 and a power of 0.95 (1 – beta error probability) to produce an actual power (decision-making criteria) for each subset. The results demonstrated that the actual power values were greater than 0.95, suggesting that the null hypothesis can be rejected, meaning that a “Loan Amount” affects whether a borrower can be charged-off. 
ANOVA test was conducted against each subset using G*Power, where the outcomes demonstrate that the actual power values are higher than 0.95, suggesting that the null hypothesis can be rejected, which means two groups offer variance differences so that a “Loan Amount” affects whether a borrower can be charged-off.
The correlation analysis was performed against each subset and produced scores of -0.005255, 0.061228, and 0.007396 per subset, where the results indicate no strong correlation between the loan amount and the status of charged-off borrowers. The correlation results are not aligned with the T-Tests, suggesting that further analysis is needed. 
\subsubsection{Hypothesis 2}
To explore the second hypothesis regarding a relationship between higher “Loan Amount” and “Charged-off,” each subset was sorted in descending order by loan amount, and the top 25\% of observations were selected for analysis. The results revealed that all actual power values were higher than 0.95, suggesting that the null hypothesis should be rejected and indicating a strong relationship between the loan amount and charged-off borrowers.  
\subsubsection{Hypothesis 3}
The third hypothesis is that the bottom 25\% of loan amounts would also show a statistical relationship with the charged-off borrowers. Each subset was sorted in descending order regarding loan amount attributed, and the bottom 25\% of observations were selected. The two-tailed T-Test (conducted by G*Power) revealed a strong relationship between the loan amount and charged-off accounts.
\section{Discussion}
The company formulated a hypothesis to explore the impact of “Loan Amount” as a dependent variable on an independent attribute referring to “Charge OFF Flag,” showing whether a borrower has repaid the loan or charged it off. To do so, LendingClub decided to conduct T-Test and ANOVA hypothesis testing and correlation analysis. The hypothesis testing revealed a statistically significant difference at p-values less than .05, which is interpreted as an indication of the impact of the loan amount on loan repayment. However, the correlation analysis produced a low score, which disagreed with the results of hypothesis testing, and the company decided to perform a more in-depth analysis to locate the source of such divergence.  
\subsection{Steps in Statistical Analysis}
Statistical analysis includes various steps, such as data exploration, hypothesis testing, and visualization, where the interpretation of results is the last step that aims to explain the results of each step (or most steps) of the analysis \cite{de2002analyzing,sarraf2022end,10.1088/978-0-7503-1793-1ch1}. In general, an explanation of statistical results often covers four main areas: a) sample size, b) metrics of central tendency, c) distribution of data, and d) hypothesis testing \cite{morgan2004spss}. 
\subsubsection{Dataset or Sample Size}
The number of observations available for statistical analysis plays a crucial role in interpreting results. This number demonstrates whether the samples (observations) can be considered representative of analyzed data \cite{goodhue2006pls}. A significant difference between statistics and machine learning exists in terms of the number of samples required for experiments, where, for example, 50 observations can represent a population for statistical analysis. A significantly larger dataset is often required for developing a machine learning model.  
\subsubsection{Measures of Central Tendency}
The mean, median, and mode of observations used for statistical analysis, along with the variance and standard deviation (i.e., measures of central tendency), reveal the central gravity of observations \cite{wilcox2003modern}. Interpreting those metrics enables practitioners to discover outliers in the observations and explore the possibility of removing them from the analysis. Unlike machine learning model development, where outliers might not impact results significantly, outliers here can affect statistical results by biasing the results towards that extreme. 
\subsubsection{Data Distribution}
Spreading data by calculating the observation variance can show how samples are distributed among a population \cite{mardia1975statistics}. Also, exploring data distribution by calculating the histogram of data can reveal the type of data distribution (i.e., normal distribution). It also indicates whether the data are skewed towards the left or right of the histogram \cite{mardia1975statistics}. Interpreting the data distribution also reveals whether the data are multimodal, where observations come from two or more distributions. Moreover, such interoperation can be used for accurate data normalization, removing outliers, and properly formulating hypotheses for future analyses or reiterations of the current analysis \cite{silverman1986density}.
\subsubsection{Hypothesis Testing}
Interpretation of hypothesis testing comprises two steps: a) exploring the logic of formulating such a hypothesis and b) exploring the results of hypothesis testing \cite{mullins2002database}. In the first step, statisticians review the reasons for forming such a hypothesis by studying documents related to the business aspects of an organization. For example, statisticians can only formulate a hypothesis for analysis because they have considered the types/amounts of loans granted as dependent variables (inputs) when predicting whether borrowers could repay \cite{emekter2015evaluating,mondal2016sensitivity}. The logic behind such a hypothesis is explored and interpreted once the data are analyzed and the results produced. The second step is to interpret the hypothesis testing results, determine whether the hypothesis is accepted or rejected, and explore the confidence interval of such interpretations \cite{berger1991interpreting}. For example, the interpretation of hypothesis testing results for types of loans and successful repayment could potentially reveal a) whether types/amounts of loans are adequate metrics for predicting risks associated with a borrower; and b) how an organization can mitigate potential risks and update their criteria for granting loans \cite{emekter2015evaluating}. 
\subsection{Limitations in Statistical Analysis}
Statistical analysis encounters various limitations that make the interpretation of results challenging. As discussed earlier, the primary challenge of statistical analysis, relative to machine learning techniques, is the number of observations required to perform analysis \cite{young2018place}. A standard practice in statistical analysis is to sample a population randomly and test hypotheses against the subset of data that can raise concerns about whether the generated subset is a true representative of data \cite{inohara2011psychological}. By contrast, training machine learning algorithms require a significant amount of data, so practitioners assume that the number of samples or observations used to train the algorithms would represent the entire population \cite{inohara2011psychological}. Another limitation in interpreting the analysis results is how to relate findings to business problems and interpret the outcomes of hypothesis testing to address business problem statements \cite{young2018place}.
\subsubsection{Small Dataset}
The size of the dataset or sample used for statistical analysis plays a crucial role in determining the extent to which the results can be generalized \cite{pasini2015artificial}. A small sample size imposes significant limitations on statistical analysis, where a small dataset serves as a somewhat unrepresentative sample of the entire population, causing different types of bias in the analysis results \cite{fong2020finding}. Also, a small dataset increases the risk that outliers in each population will negatively impact measures of central tendency that have been calculated based on samples out of distribution. In addition to the problem of outliers discussed earlier, a small dataset makes splitting data into training and testing highly challenging. Although statistical analysis methods employ all samples provided to implement models based on hypothesis testing, practitioners in the field often use unseen data to validate hypothesis testing results \cite{fong2020finding,pasini2015artificial}. Another issue caused by a small sample size is an unpredicted increase in measurement errors where the error metrics used to evaluate the models produce highly varying results. To overcome the limitations imposed by a small dataset, the primary practice is to randomly shuffle the dataset and generate several subsets of data, repeating statistical analysis to ensure the results converge \cite{kvesic2012risk}.
\subsubsection{Cause and Effect}
One of the challenges in interpreting statistical results relates to inconsistency between the hypotheses formulated and the outcomes of testing methods. Practitioners interpreting the statistical results might notice that the results are misaligned with the logic of hypothesis tests \cite{doggett2004statistical}. In such ambiguous circumstances, discovering the cause and effect in statistical analysis results conducted on specific business use cases is challenging since the interpretation disagrees with the predefined scenario \cite{doggett2004statistical}. This issue can arise when the hypothesis testing design does not cover the useful parameters in testing or when less powerful features and attributes in data are used for hypothesis testing \cite{laland2011cause}. It sometimes happens that practitioners or business teams helping design such statistical analysis misinterpret the results or overlook some findings and/or implications \cite{doggett2004statistical,laland2011cause}. Another source of issues includes a low confidence interval level and results lacking statistical significance \cite{doggett2004statistical}. 
\subsubsection{Divergence of Results Obtained from Various Methods}
A common challenge in interpreting statistical analysis results occurs when the results obtained from various techniques diverge \cite{read2012goodness}. It is a widespread practice that statisticians design a statistical analysis using multiple techniques, such as T-Test, ANOVA, or regression, to explore whether the results produced by these techniques align. An agreement between the results from different methods enables an organization to interpret analytical results clearly and make firm recommendations. However, the research shows that hypothesis testing and other methods, such as correlation analysis or machine learning, sometimes produce different results, contrasting with other methods\cite{read2012goodness}. Such an issue indicates that a systematic problem might exist in preparing samples or conducting hypothesis testing. The solution for this type of problem is offered case by case, where practitioners more familiar with the organization’s business scope can suggest methods that produce results closer to the problem statement.
\subsection{Business Statistical Analysis and Interpretation}
Business statistics, which include various types of analysis, focus on statistical methodologies aligned with an organization’s business scope to improve the decision-making process, mitigate risks to the organization, and increase revenue \cite{sun2022improving}. Interpretation of such analysis is crucial to the organization, and the process is expected to go beyond that of a simple report or presentation. The areas covered by business statistics include a) customer behavior prediction and trend extraction; b) data exploration, hypothesis testing, and interpretation, such as extensive visualization; c) enhancing business performance from various angles; and d) improving decision-making processes \cite{sun2022improving}. To achieve such targets, business data analysts understand their organization’s business objectives and explore data and results. Also, the root cause analysis is performed to extract in-depth technical insights regarding the organization’s vulnerabilities, enabling the organization to inform its decision-making process \cite{sun2022improving}. 
\subsection{Reflection on the Statistical Analysis Process}
The findings from the initial statistical application enable the company to redesign the statistical analysis processes to concentrate on those attributes that more substantially impact their business. Feature engineering—a systematic methodology—is necessary to reveal the relationships between dependent attributes and target variables \cite{nargesian2017learning}. Also, the company aims to explore other features highly correlated with potential target variables from the business perspective but uncorrelated with other dependent attributes \cite{kotusev2019enterprise}. 
\subsubsection{Potential Improvement}
The process of statistical analysis at LendingClub requires several changes to better serve the company’s business needs. The primary targets are to enhance the process of issuing loans, such as the duration of the loan approval process, and to mitigate financial risks to the company by offering borrowers a data-driven loan amount. LendingClub is to apply such changes to the statistical analysis and decision-making process by employing big data infrastructure for advanced multi-model data collection and analytics. In the first step, the company needs a plan demonstrating how to onboard new technology and its costs. The second step includes a broader statistical analysis, such as hypothesis testing, and uses the current data to assess whether specific statistical applications could broadly improve the company’s performance. In the third step, LendingClub conducts research and recruits a third party to develop the required infrastructure.
\subsubsection{Required Infrastructure}
\begin{figure}
	\centering{\includegraphics[width=\textwidth,height=\textheight,keepaspectratio]{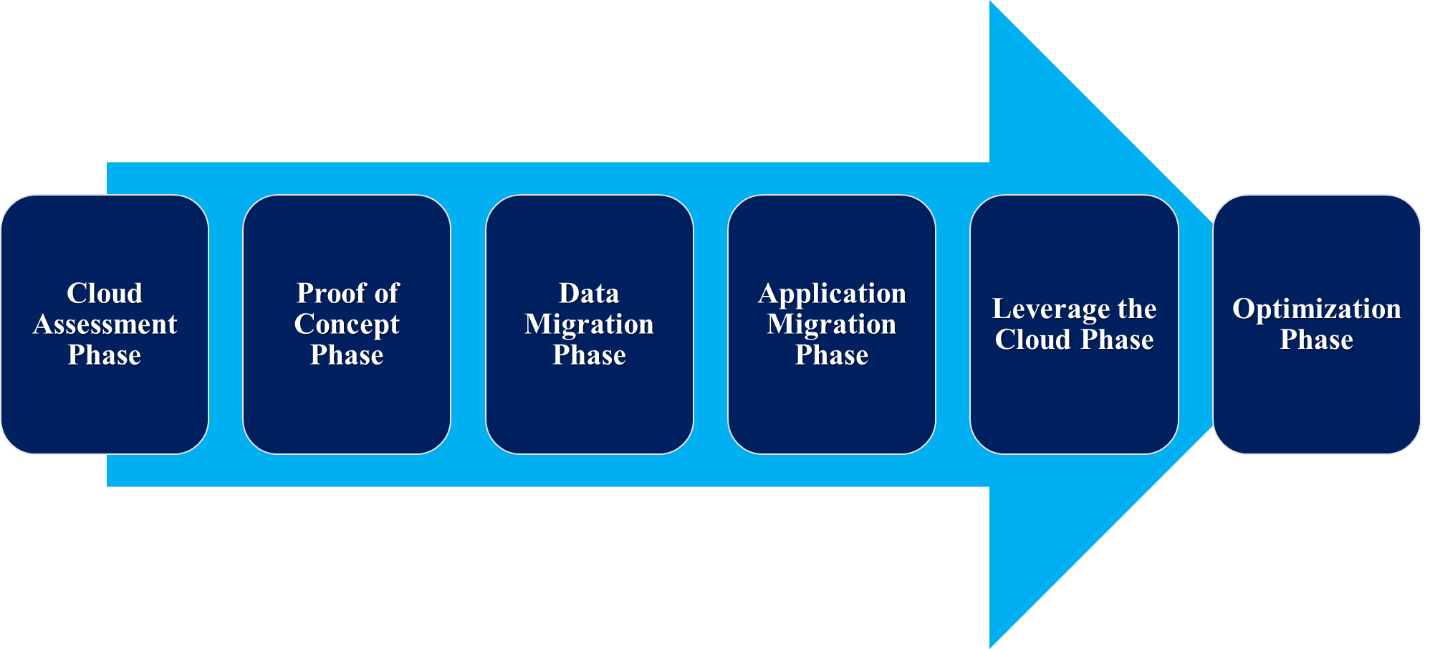}}
	\caption{Steps for migrating data pipeline to a cloud environment}
	\label{fig:fig2}
\end{figure}

Onboarding a large-scale system, such as an enabled big data analytics platform, is a significant change to LendingClub, where modifications have been performed to everything from databases to reporting systems. The first stage is to decide whether LendingClub would adopt a big data platform to the current system or entirely migrate to the new model. This decision allows the stakeholder to estimate the cost of a big data platform and start planning. Although the cost of system adaptation or migration to the big data platform requires detailed information, the migration to a cloud environment, for example, offering various big data services, would be a potential expansion of LendingClub’s analytics in the future. Figure \ref{fig:fig2}   illustrates the proposed steps for migrating the LendingClub data collection and analytics pipeline to a cloud-based environment that offers big data services such as Amazon Web Services (AWS) \cite{al2019big,mullins2002database}. These steps consist of a) cloud assessment, b) proof of concept, c) data migration, d) application migration, e) leverage of the cloud, and f) optimization. 
\subsection{Proposed Large-Scale Plan}
\begin{figure}
	\centering{\includegraphics[width=\textwidth,height=\textheight,keepaspectratio]{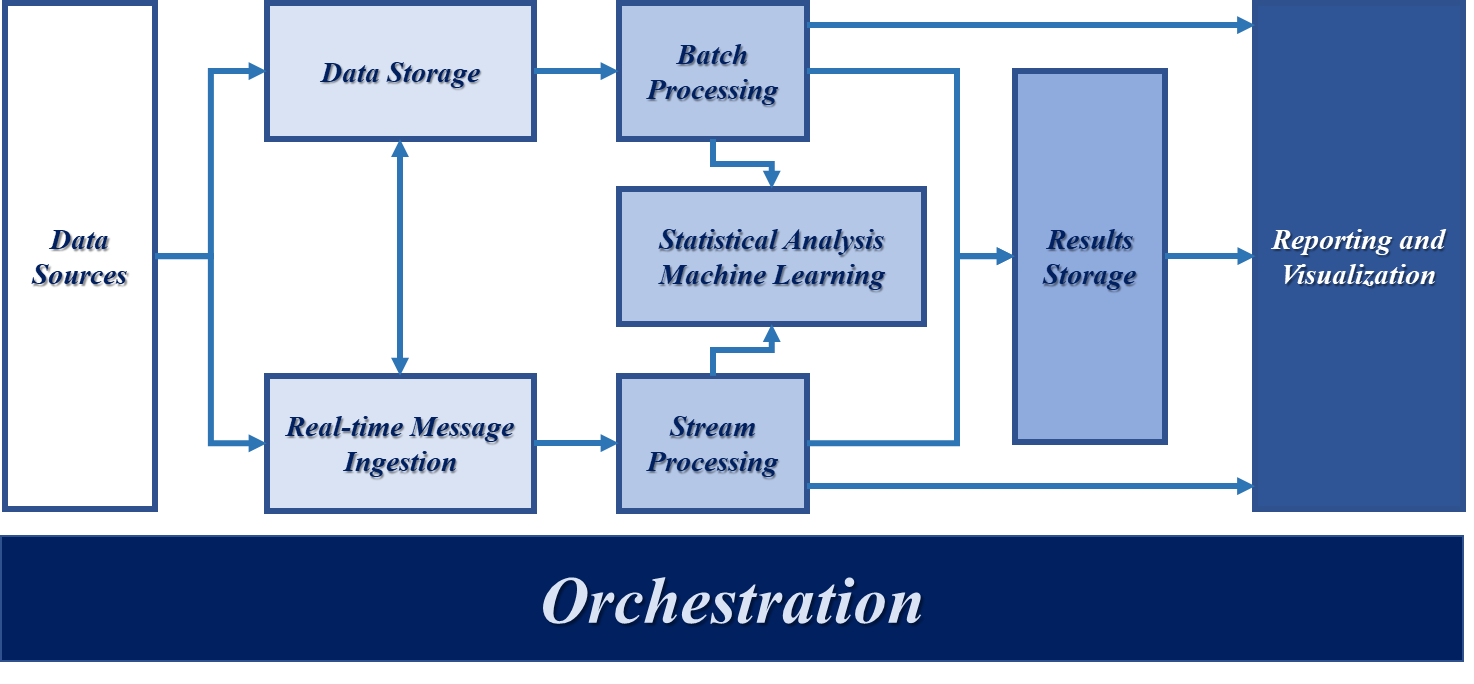}}
	\caption{The architecture of data collection and analysis pipeline}
	\label{fig:fig3}
\end{figure}

The large-scale plan to enhance the current statistical analysis pipeline consists of two primary phases: a) designing and implementing an end-to-end data collection and processing pipeline that offers big data analytics, and b) increasing the number and quality of features \cite{lee2019cloud}. The current data collection pipeline collects data from various sources, and no broadly systematic methodology is employed to acquire such data. Gathering data from different providers (in-house or third-party) involves an extensive preprocessing pipeline, which might remove many observations to prepare a consistent dataset. 
\\
The proposed pipeline illustrated in Figure \ref{fig:fig3} offers various capabilities, including big data collection and data stream processing. The first component of the architecture is a user interface that enables it to receive data from external sources where the data could either be stored in a multi-model database or be in the form of real-time messaging input into an allocated database. The collected data can be transferred between data storage and real-time messaging place holders, which offers big data capabilities to host structured and unstructured data. The next architecture layer includes enabled big data processing components for batch processing, which oversees data preparation and preprocessing for further analysis \cite{sarraf2019mcadnnet}. 

A similar component—the stream processing unit—prepares and preprocesses data streams for real-time analysis and applications. The preprocessed data are sent to the next component of the architecture, which encompasses the statistical analysis and machine learning methods, where such a block is considered the brain that orchestrates the data analytics. Statistical analysis or machine learning outcomes are stored in a “results database.” The last layer of this orchestration is the user interface block, which enables practitioners in the organization to generate reports with visualizations that can be provided to leadership for decision-making purposes. An extra capability in the new architecture is scheduling automatic training machine learning models or performing statistical analysis. 
\\
The second phase of the new data analytics platform aims to enhance the quality of feature selection, which concentrates on those attributes that contribute most to target variables. Quarter-based statistical analysis and feature engineering demonstrate what features should be collected with higher resolution. The advantage of using targeted data collection through particular data attributes is to reduce the cost of on-demand infrastructure by reducing the load on the architecture servers and analytical blocks. However, the main disadvantage of employing such a step is that it decreases the amount of data that can be collected, which might harm statistical analysis or predictive model development. Therefore, the organization must weigh the cost of massive data streaming and collection against the impact of selective data collection.
\section{Conclusions}
Statistical applications enable enterprises to establish a data-driven business plan that provides clear objectives to enhance the enterprise’s performance, revenue, and risk management. This work summarized a strategic plan informed by an already performed analysis for LendingClub – a financial company – that grants various forms. The statistical results showed that different logic could be extracted from currently collected data. Such results enabled LendingClub to improve its business scope and to encourage the company to onboard a big data platform. The plan recommended exploring employing enhanced feature engineering capabilities to acquire enormous data per year and develop predictive models to increase the company’s revenue and lessen potential risks. LendingClub’s plan also seeks to utilize artificial intelligence and machine learning technologies to implement robust models aligned with the company’s business scopes.

\bibliographystyle{unsrtnat}
\bibliography{references}  






\end{document}